\documentclass{article}

\usepackage{arxiv}

\usepackage[utf8]{inputenc} 
\usepackage[T1]{fontenc}    
\usepackage{hyperref}       
\usepackage{url}            
\usepackage{booktabs}       
\usepackage{amsfonts}       
\usepackage{nicefrac}       
\usepackage{microtype}      
\usepackage{lipsum}
\usepackage{graphicx}
\graphicspath{ {./figs/} }

\usepackage{amsfonts}
\usepackage{amsmath}
\usepackage{amssymb}
\usepackage{amsthm}
\usepackage{cleveref}
\usepackage{siunitx}
\usepackage{listings}

\crefname{lstlisting}{Listing}{Listings}
\Crefname{lstlisting}{Listing}{Listings}

\usepackage{xcolor}

\definecolor{hgf-blue}{RGB}{0,90,160}
\definecolor{hgf-green}{RGB}{140,180,35}
\definecolor{hgf-gray}{RGB}{90,105,110}
\definecolor{hgf-purple}{RGB}{160,35,90}
\definecolor{hgf-red}{RGB}{210,50,100}
\definecolor{hgf-orange}{RGB}{240,120,30}
\definecolor{hgf-yellow}{RGB}{255,210,40}
\definecolor{hgf-turqoise}{RGB}{80,200,170}
\definecolor{hgf-dark-green}{RGB}{50,100,105}

\title{Accelerating Neural Network Training with \\
Distributed Asynchronous and Selective Optimization (DASO)}

\author{
 Daniel Coquelin \\
   Steinbuch Centre for Computing (SCC) \\
   Karlsruhe Institute of Technology (KIT) \\
   Karlsruhe, Germany \\
  \texttt{daniel.coquelin@kit.edu} \\
  \And
   Charlotte Debus \\
   Steinbuch Centre for Computing (SCC) \\
   Karlsruhe Institute of Technology (KIT) \\
   Karlsruhe, Germany \\
   \texttt{charlotte.debus@kit.edu} \\
  \And
   Markus Götz \\
   Steinbuch Centre for Computing (SCC) \\
   Karlsruhe Institute of Technology (KIT) \\
   Karlsruhe, Germany \\
   \texttt{markus.goetz@kit.edu} \\
  \And
   Fabrice von der Lehr \\
   Institute for Software Technology (SC) \\
   German Aerospace Center (DLF) \\
   Cologne, Germany \\
   \texttt{fabrice.lehr@dlr.de} \\
  \And
   James Kahn \\
   Steinbuch Centre for Computing (SCC) \\
   Karlsruhe Institute of Technology (KIT) \\
   Karlsruhe, Germany \\
  \texttt{james.kahn@kit.edu} \\
  \And
   Martin Siggel \\
   Institute for Software Technology (SC) \\
   German Aerospace Center (DLF) \\
   Cologne, Germany \\
   \texttt{martin.siggle@dlr.de} \\
  \And
   Achim Streit \\
   Steinbuch Centre for Computing (SCC) \\
   Karlsruhe Institute of Technology (KIT) \\
   Karlsruhe, Germany \\
   \texttt{achim.streit@kit.edu} \\
}

\begin{document}
\maketitle

\begin{abstract}
With increasing data and model complexities, the time required to train neural networks has become prohibitively large. To address the exponential rise in training time, users are turning to data parallel neural networks (DPNN) to utilize large-scale distributed resources on computer clusters. Current DPNN approaches implement the network parameter updates by synchronizing and averaging gradients across all processes with blocking communication operations. This synchronization is the central algorithmic bottleneck. To combat this, we introduce the Distributed Asynchronous and Selective Optimization (DASO) method which leverages multi-GPU compute node architectures to accelerate network training. DASO uses a hierarchical and asynchronous communication scheme comprised of node-local and global networks while adjusting the global synchronization rate during the learning process. We show that DASO yields a reduction in training time of up to 34\% on classical and state-of-the-art networks, as compared to other existing data parallel training methods.



\end{abstract}



\maketitle

\section{Introduction}
\label{sec:intro}

Recent advances in deep learning have thrived under the theme "bigger is better". Modern neural networks yield super-human performance on problems such as image classification and semantic segmentation by introducing higher model complexity, for example more layers, inter- and intra-layer connections~\cite{he2016resnet,vaswani2017attention}. However, the training of large networks also requires large datasets. As the sizes of models and datasets increases, so do the computational resources required. In other words, today's deep learning tasks are limited by the hardware and computing time available. In response, parallel training methods have been developed to enable the concurrent use of multiple (distributed) hardware devices.

In general, there are two approaches to parallel training~\cite{ben2019demystifying}: model parallelism and data parallelism. The model parallel approach distributes the network across multiple computing devices, for example two GPUs with half of the network each. In the data parallel approach, each available computing device trains an identical copy of the network.

Data parallel neural networks (DPNNs) have been used on various architectures and data types to achieve state-of-the-art results~\cite{yamazaki2019accelerated, tao2020hierarchical}. Each model instance in a DPNN performs a forward-backward pass individually over a unique portion of the data, after which the parameters of all networks are synchronized using a global collective operation. This can be effectively viewed as a batch distributed across the devices, i.e. a distributed batch. Traditionally, the synchronization of network parameters is a blocking, averaging operation ~\cite{ben2019demystifying}. This collective blocking operation comprises an inherent bottleneck.

Using non-blocking operations can provide relief as the next forward-backward step can begin while communication is ongoing. However, as global parameter updates are running asynchronously, parameters found by individual network instances are always slightly out-of-date. Out-of-date parameters can also be referred to as stale. 

Although computing devices can take many forms, GPUs are currently the most efficient and powerful for training neural networks. Therefore, we will refer to computing devices as GPUs throughout this paper.

Averaging network parameters across multiple instances, traditionally referred to as mini-batch optimization, is only an approximation of the true gradients that would be calculated over the unified batch with batch optimization. Moreover, the standard communication structure communicates with each GPUs individually to synchronize network parameters.
This neglects the structure of most computer clusters, where multiple GPUs are grouped on computing nodes with significantly faster node-local connections as compared to cross-node communication. 

Large multi-node DPNNs can instead be divided into node-local DPNNs which are themselves members of a global DPNN. This hierarchical approach would significantly reduce the communication overhead, as less data is sent between nodes. Furthermore, what if global parameter synchronization did not occur after every batch and instead the average was calculated asynchronously every $B$\textsuperscript{th} batch?

To this end, we present our key contribution: the distributed selective and asynchronous optimization (DASO) method. DASO performs communication for network parameter updates in a hierarchical manner: on the node-local level, in the form of GPU-to-GPU operations, and on the global level, where computing nodes are treated as individual entities. This approach allows DASO to perform the time-expensive global synchronization after multiple batches instead of after every forward-backward pass, thus leveraging the potential of acceleration via parallel computation on modern computer clusters.

The remainder of this paper is organized as follows. In \Cref{sec:related} we will discuss relevant work previously done in the area of data parallel model training. \Cref{sec:optimizer-bkg} introduces the concept of selective distributed asynchronous optimization, followed by performance evaluations on the tasks of image classification and semantic segmentation in \Cref{sec:experiments}. Our results are summarized and discussed in \Cref{sec:conclusion}, which also gives an outlook towards further improvement and application of the method.

\section{Related Work}
\label{sec:related}

Data parallel neural networks are the go-to option for accelerating training on large datasets. In DPNNs, each local network is optimized locally, e.g using SGD, before the optimization results are synchronized with all other networks. 
The most straightforward approach to global synchronization is a collective blocking, average operation after every forward-backward step. This inherently limits the speed of the data parallel training.

Recently, advancements have been made in accelerating the synchronization process by starting the communication of gradient updates while the backward pass is ongoing, with one reporting training times of only \SI{74.7}{\second} on the ImageNet data set~\cite{yamazaki2019accelerated}. However, this is a tailored approach which does not generalize, as it is highly optimized for a specific network and requires considerable tuning.
	
Several works have investigated the use of asynchronous SGD (ASGD)~\cite{sa2017asgd, lian2018asynchronous, zhang2013asgd}, which updates the parameters whenever a network finishes a backward pass. Each network retrieves the current model parameters from a parameter server before performing a forward-backward pass. After finishing the backward step, the network sends its updated parameters back to the server, which determines the new global parameters using the updates from all processes. However, if a network is still computing the forward-backward pass when the parameter server is updated, the network's current parameters are outdated. The subsequently found gradients are referred to as stale. 
Stale gradients can be leveraged to approximate accurate network parameters, and ASGD has been shown to yield consistent convergence~\cite{zhang2016stalenessaware}. Recent attempts at accelerating ASGD have been made using individual network optimizers for a warm-up phase and delayed updates to the parameter server~\cite{bogoychev2018accelerating}.

PyTorch~\cite{paszke2019pytorch} and TensorFlow~\cite{abadi2015tensorflow} are currently the largest machine learning frameworks. Both offer options for traditional data parallel training. For large systems, a global communication protocol, such as MPI~\cite{mpi2015mpi}, is often required to leverage specialized inter-node connections. Recently, there have been many advancements in the optimization of the global parameter synchronization operation by using MPI with multiple network topologies~\cite{ueno2019mpi-hierarchical, mikami2018mpi-tort}. These approaches have shown promising results, but remain centered around the idea of a global synchronization for each forward-backward pass.

Currently, the most popular MPI-enabled DPNN framework is Horovod~\cite{sergeev2018horovod}. To reduce the size of data sent via the communication network, Horovod uses tensor fusion, or grouping parameters together to be communicated in a larger chunk of data, and data compression. The data compression in Horovod is frequently done by casting the network parameters into 16-bit floating-point format. 

\section{Distributed Asynchronous and Selective Optimization (DASO)}
\label{sec:optimizer-bkg}

The common approach to training DPNNs is to perform a forward-backward pass on each network instance with one portion of the distributed batch, then synchronize the network parameters via a global averaging operation. The averaging of gradients is only an approximation of the true gradients that would be calculated for the entire batch when processes on a single GPU. This approximation is made under the assumption that each portion of the distributed batch is independent and identically distributed (iid)~\cite{clauset2011brief}.

Under the iid assumption, another approximation can be made: the average parameters of a subset of networks are not significantly different than the average parameters of the complete set of networks. Recalling that modern HPC clusters have different inter- and intra-node communication capabilities (with different bandwidths and latencies), we can utilize this approximation to reduce the communication needed for parallel training, thereby alleviating the intrinsic bottleneck of blocking synchronizations.

We therefore propose the Distributed Asynchronous and Selective Optimization (DASO) method. Instead of a uniform communications network across multiple multi-GPU nodes, DASO uses a hierarchical network model with node-local networks and a global network. 

\begin{figure}
    \centering
    \includegraphics[width=0.95\linewidth]{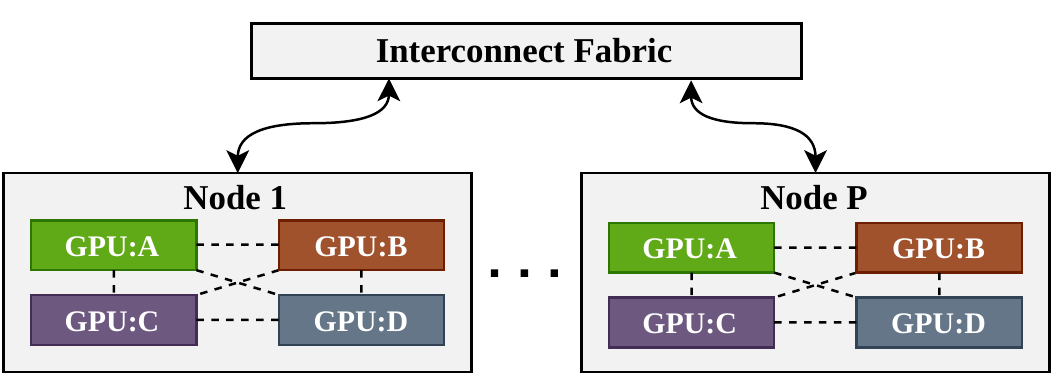}
    \caption{An overview of a common node-based computer cluster with P nodes and four GPUs per node. GPU colors represent \emph{group} membership. The dashed lines indicate GPU-to-GPU communication channels.}
\label{fig:mpi-groups}
\end{figure}

The global network spans all GPUs on all nodes, while the node-local networks are composed of the GPUs on each individual node. The global network is divided into multiple \emph{groups}, with each \emph{group} containing a single GPU from every node. Global communication takes place exclusively within a \emph{group}, i.e. only \emph{group} members exchange data, while members of other \emph{groups} do not participate. Communication between the node-local GPUs is then handled by the local network, which benefits from high-speed GPU-to-GPU interconnects and optimized communication packages (e.g. NCCL~\cite{li2020nccl}).
Under the assumption that the cluster node configurations are homogeneous, DASO creates \emph{groups} between GPUs with the same local identifier as is shown in \Cref{fig:mpi-groups}. With this approach, inter-node communication can be reduced by a factor equal to the minimum number of GPUs per node.

\begin{figure}[ht]
    \centering
    \includegraphics[width=0.95\linewidth]{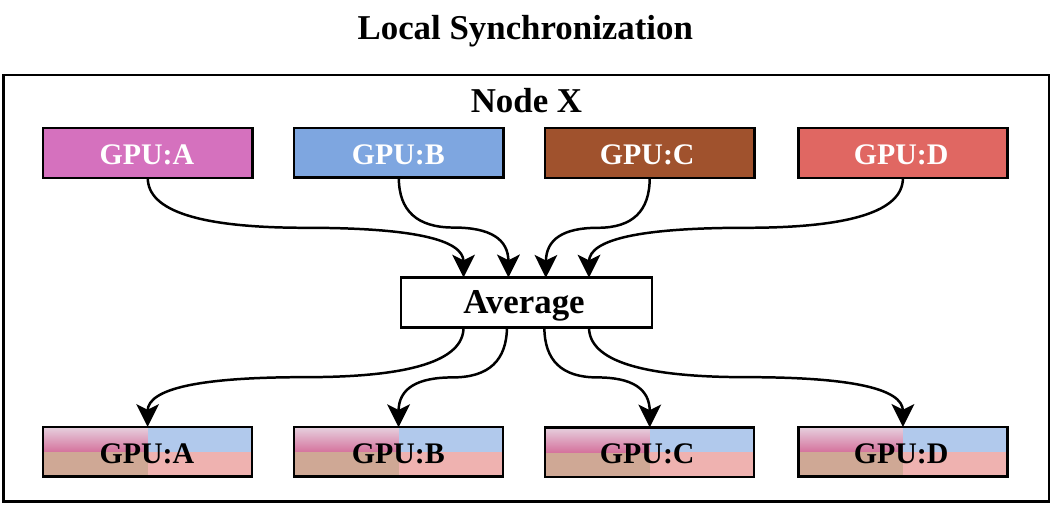}
    \caption{Schematic of the local synchronization step for a single node with four GPUs. The gradients from each GPU are averaged, then each GPU's gradients are set to the result.}
\label{fig:local-sync}
\end{figure}

\begin{figure}
    \centering
    \includegraphics[width=0.95\linewidth]{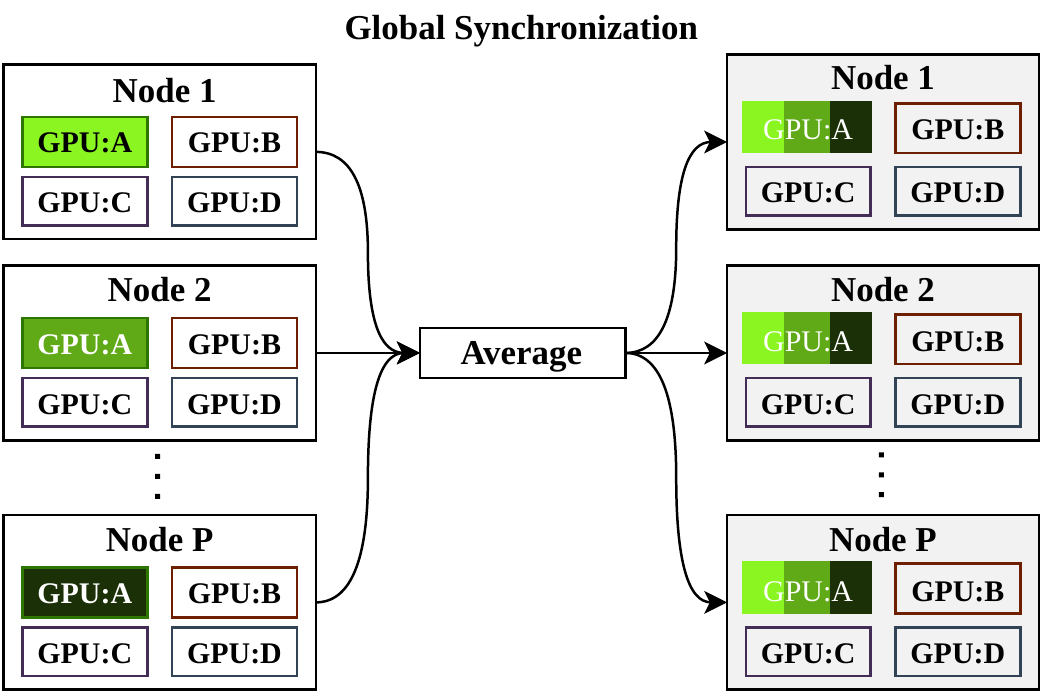}
    \caption{Schematic of the global synchronization step performed by the global communication \emph{group} consisting of GPU:A on each node. The network parameters are averaged by each GPU in the \emph{group}, and the network parameters of each \emph{group} member are set to the result.}
\label{fig:global-sync}
\end{figure}

\begin{figure}
    \centering
    \includegraphics[width=0.95\linewidth]{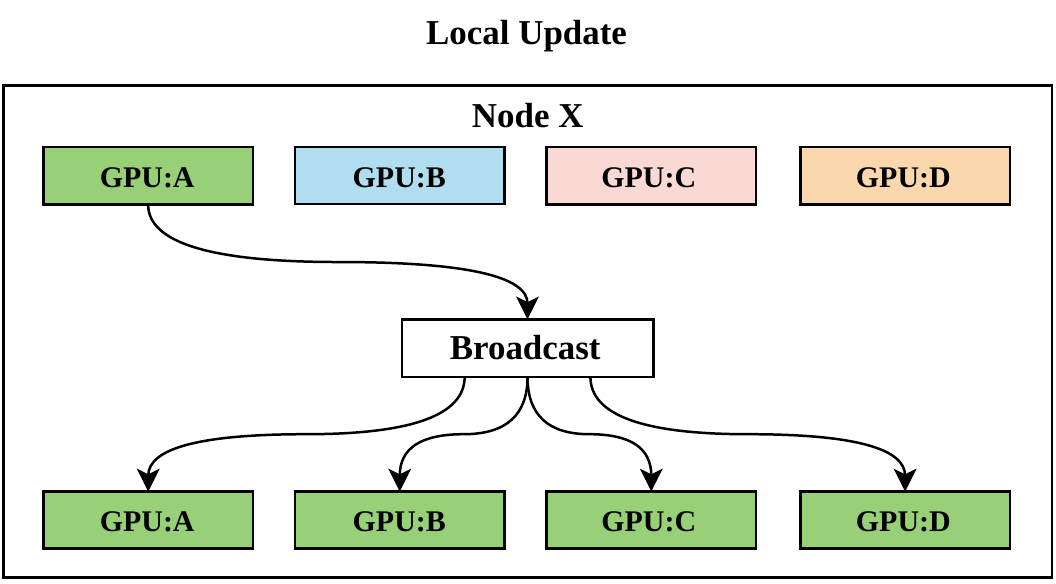}\hfill
    \caption{Schematic of the local update step to be performed after the global synchronization step shown in \Cref{fig:global-sync}. The group member responsible for the global communication, in this case GPU:A, sends its network parameters to all other node-local GPUs, which replace the old parameters on those GPUs.}
\label{fig:local-bcast}
\end{figure}

DASO utilizes a multi-step synchronization. Local synchronization (\Cref{fig:local-sync}) occurs after each batch and uses the node-local network to do gradient-averaging between the local GPUs. Global synchronization (\Cref{fig:global-sync}) occurs after one or more local synchronizations, in which the network parameters of all members of a single global \emph{group} are shared and averaged. Following every global synchronization, a local update step broadcasts averaged parameters from the local \emph{group} member to all other node-local GPUs (\Cref{fig:local-bcast}). The role of global synchronization rotates between \emph{groups} to overlap communication and computation.
Global synchronization can be performed in a blocking or non-blocking manner. In the blocking case, all synchronization steps are performed after each batch. To reduce the amount of data transferred, parameters are cast to a 16-bit datatype representation during buffer packaging. This operation does not effect convergence, as shown by~\cite{alistarh2017qsgd}. Once received, the parameters are cast back to their original datatype.
In the non-blocking case, the next forward-backward pass is started after the parameters are sent but before they are received. 
Datatype casting is not beneficial in this scenario, as it delays the start of parameter communications. Each neural network will conduct $B$ forward-backward passes complete with local synchronization before the \emph{group} members receive the sent parameters. Hence, the updates from the global communication step are outdated upon their arrival. To compensate for this, a weighted average of the stale global parameters and the current local parameters is calculated as follows:
\begin{align}
\label{eq:weighted-avg-txt}
    x_{t+S} &= \frac{2Sx^l_{t+S-1} + \sum^{P}_{i=1} x^{i}_{t}}{2S + P} 
\end{align}
where $x^l_{t+S}$ is the model state on GPU $l$ after $S$ batches to wait after starting batch $t$ for the global synchronization data, $x^i_{t+1}$ is the globally exchanged model states, and $P$ is the number of GPUs in the global network. The weighting of the local parameters was found experimentally. A detailed explanation of \Cref{eq:weighted-avg-txt} and its validity is provided in the supplementary material.

Training of a network with the DASO method can be divided into three key phases: warm-up, cycling, and cool-down. The warm-up and cool-down phases utilize blocking global synchronizations, while the cycling phase uses non-blocking global synchronizations. Given a fixed number of total epochs, warm-up and cool-down phases occur for a set number of epochs at the beginning and end of training respectively. The warm-up phase is used to quickly move away from the randomly initialized parameters and prepare for the cycling phase. The cool-down phase is intended to reduce the slight errors which can arise due to the slight deviation from the iid assumption for individual batches.

\begin{figure}[ht]
    \centering
    \includegraphics[width=0.5\linewidth]{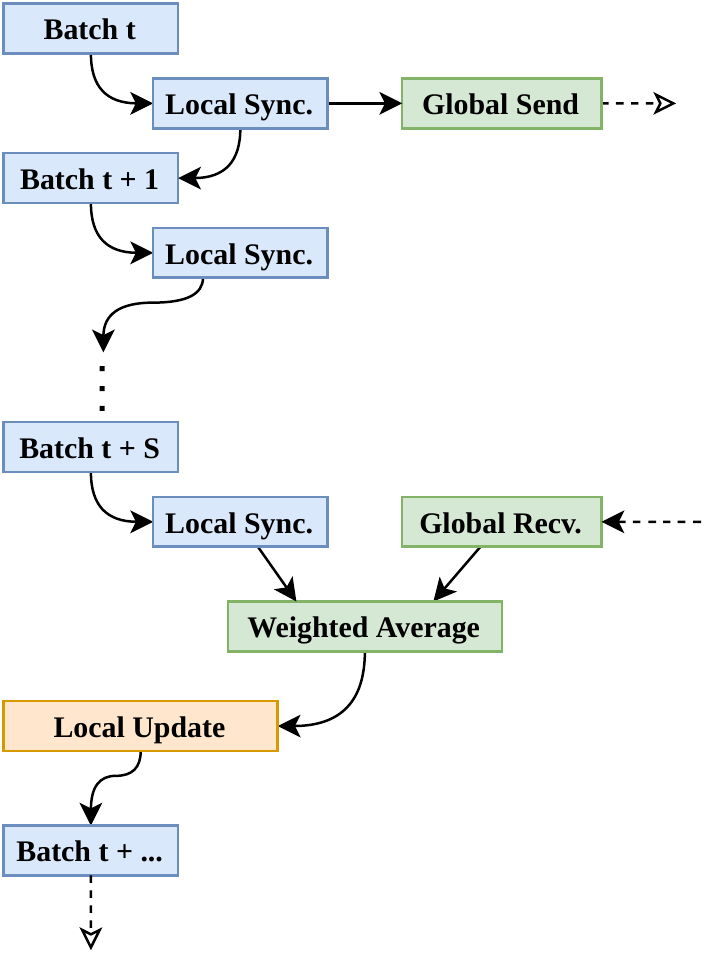}
    \caption{Process flow diagram of the synchronization steps during the cycling phase where t is the batch number and S is the batches to wait before global synchronization. The weighted average is calculated as shown in \Cref{eq:weighted-avg-txt}}
\label{fig:skip-batches-flow}
\end{figure}

In the cycling phase, the number of forward-backward passes between global synchronizations ($B$) and the number of batches to wait for global synchronization data ($W$) are varied. $B$ is specified manually upon initialization. For $W$, an initial value of $B/4$ was found empirically to perform best. Each time the training loss plateaus, $B$ and $W$ are reduced by a factor of two, down to a minimum of one. When $B, W = 1$ and the loss has plateaued, both are reset to their initial values and the process is repeated until the cool-down phase. The synchronization steps in the cycling phase are schematically shown in \Cref{fig:skip-batches-flow}.

\subsection{Current Implementation}
\label{seg:daso-imp}

A DASO proof-of-concept is currently implemented in the HeAT framework~\cite{goetz2020heat} for usage with PyTorch networks. HeAT is an open-source Python framework for distributed and GPU-accelerated data analytics which offers both low level array computations as well as assorted higher-level machine learning algorithms. The local networks utilize PyTorch's \texttt{DistributedDataParallel} class and distributed package~\cite{li2020pytorch}. The global communication network utilizes HeAT's MPI backend, which handles the automatic communication of PyTorch Tensors. The global \emph{groups} are implemented as MPI groups.

To use this implementation of DASO to train an existing PyTorch network, only four additional functions need to be called and the data loaders need to be modified to distribute the data between all GPUs\footnote{The data loaders need only know how many GPUs exist and what their global rank is.}. The function calls are illustrated in \Cref{lst:daso-code}. First, the node-local PyTorch processes are created, which will be utilized during the local synchronization step. Next, the optimizer instance, i.e. DASO, is created with a PyTorch node-local optimizer (e.g. SGD) and the number of epochs for training is specified. The DASO instance will find the aforementioned PyTorch processes automatically. 
\vspace{2cm}

\begin{lstlisting}[
    caption={Simplified training script demonstrating the usage of DASO in HeAT for a PyTorch neural network (\texttt{net}) and PyTorch optimizer (\texttt{optimizer}).\\},
    label={lst:daso-code}
]
import heat as ht
import torch
...
# create PyTorch distributed group
world_size = ht.MPI_WORLD.size
rank = ht.MPI_WORLD.rank
local_rank = rank % num_local_gpus
torch.distributed.init_process_group(
    backend="nccl", 
    rank=local_rank, 
    world_size=world_size
)
...
# the DASO optimizer is created
daso_optimizer = ht.optim.DASO(
    local_optimizer=optimizer,
    total_epochs=num_epochs
)
...
# the hierarchical network is created
ht_model = ht.nn.DataParallelMultiGPU(
    net, 
    daso_optimizer
)
\end{lstlisting}

\section{Performance Evaluation}
\label{sec:experiments}

We evaluate the DASO method on two common examples of data-intensive neural network challenges: a) image classification and b) semantic segmentation. For image classification, we trained ResNet-50~\cite{he2016resnet} on the ImageNet-2012~\cite{deng2009imagenet} dataset. This can be considered a standard benchmark for machine learning, since pre-trained ResNet-50 networks are the backbone of many computer vision pipelines~\cite{wu2019detectron2}. For semantic segmentation, we trained a state-of-the-art hierarchical multi-scale attention network~\cite{tao2020hierarchical} on the CityScapes~\cite{cordts2016cityscapes} dataset. 

All experiments were conducted on the JUWELS Booster at the J\"{u}lich Supercomputing Center~\cite{krause2019juwels}. This centers' HPC cluster has 936 GPU nodes each with two AMD EPYC Rome CPUs and four NVIDIA A100 GPUs, connected via an NVIDIA Mellanox HDR InfiniBand interconnect fabric. The following software versions were used: CUDA 11.0, ParaStationMPI 5.4.7-1-mt, Python 3.8.5, PyTorch 1.7.1+cu110, Horovod 0.21.1, and NCCL 2.8.3-1. The JUWELS Booster provides a CUDA-aware MPI implementation, meaning that GPUs can communicate directly with other GPUs.

We compared DASO to Horovod, as this is currently the most popular choice for MPI-based parallel training of neural networks on computer clusters. We elected not to compare with PyTorch's distributed package as it utilizes a similar approach to Horovod, namely compression and bucketing. Comparisons are done with respect to training time and accuracy. 

Relevant network hyperparameters remain consistent for DASO and Horovod for each experiment.
All tested networks use a learning rate scheduler. When the training loss plateaus, i.e. the training loss is not decreasing by more than a set percentage threshold, the scheduler decreases the learning rate by a set factor. Settings of the scheduler, as well as for the local optimizer settings, were set to be identical for both DASO and Horovod for each use-case. With respect to message packaging, Horovod was configured to use floating point 16 compression while DASO compresses to brain floating point 16.
The batch size of training is fixed for each GPU in all experiments. Therefore, the combined distributed batch size increases by the number of GPUs times the local batch size. DASO's maximum number of batches between global synchronizations was set to four for both experiments.

\subsection{Image Classification -- ImageNet}
\label{seg:imagenet}

This experiment was conducted using the ResNet-50 architecture on the ImageNet dataset~\cite{deng2009imagenet}.
For this experiment, the ImageNet-2012 is a large dataset containing 1.2 million labeled images.
We evaluate classification quality using top-1 accuracy, i.e. the accuracy with which the model predicts the image labels correctly with a single attempt. For training ResNet-50 on the ImageNet dataset, we consider a 75\% top-1 accuracy to be a successful training.

File loading from disk and preprocessing were done using DALI~\cite{nvidia2021dali}. Training was conducted using cross entropy loss and SGD with a momentum of 0.9 and weight decay of 0.0001 for 90 epochs with a learning rate warm-up phase of five epochs. These values were adapted from PyTorch's example training script for ResNet-50 on ImageNet. The maximum learning rate is scaled with the number of global processes. The learning rate decays by a factor of 0.5 when the training cross entropy loss is stable for 5 epochs. 

Training was conducted on 4, 8, 16, 32, and 64 nodes, which equals 16, 32, 64, 128, and 256 GPUs, respectively. This corresponds to traditional strong scaling experiments for parallel algorithms, where an increase in nodes should ideally result in a proportional reduction in time. 

\begin{figure}
\centering
\begin{minipage}{.475\textwidth}
  \centering
  \includegraphics[width=\linewidth]{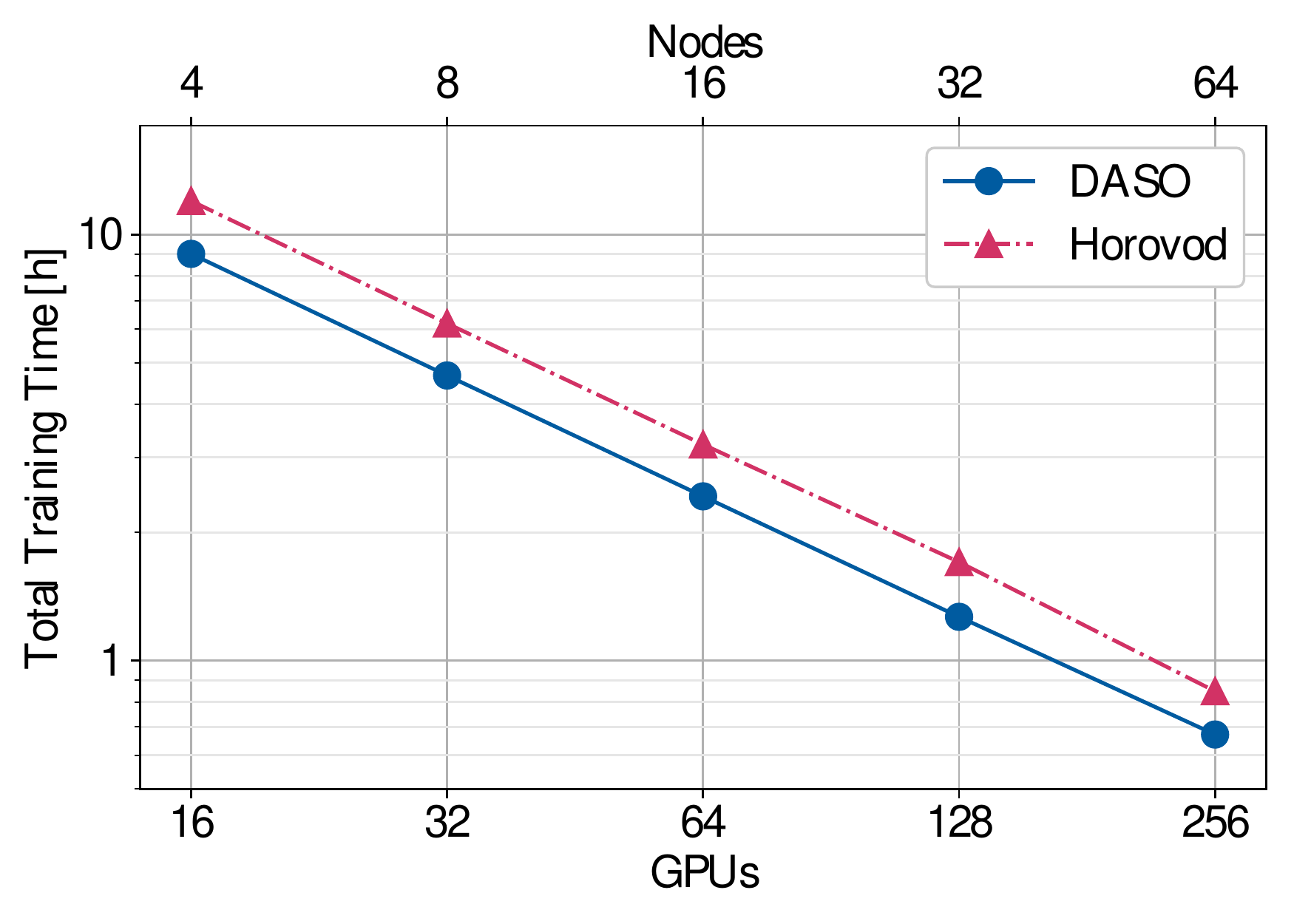} \hfill
    \caption{ResNet-50 training time on the ImageNet dataset with DASO and Horovod for increasing node counts. Each node has 4 GPUs.}
\label{fig:img-scaling}
\end{minipage}
\hspace{.01\textwidth}
\begin{minipage}{.475\textwidth}
  \centering
  \includegraphics[width=\linewidth]{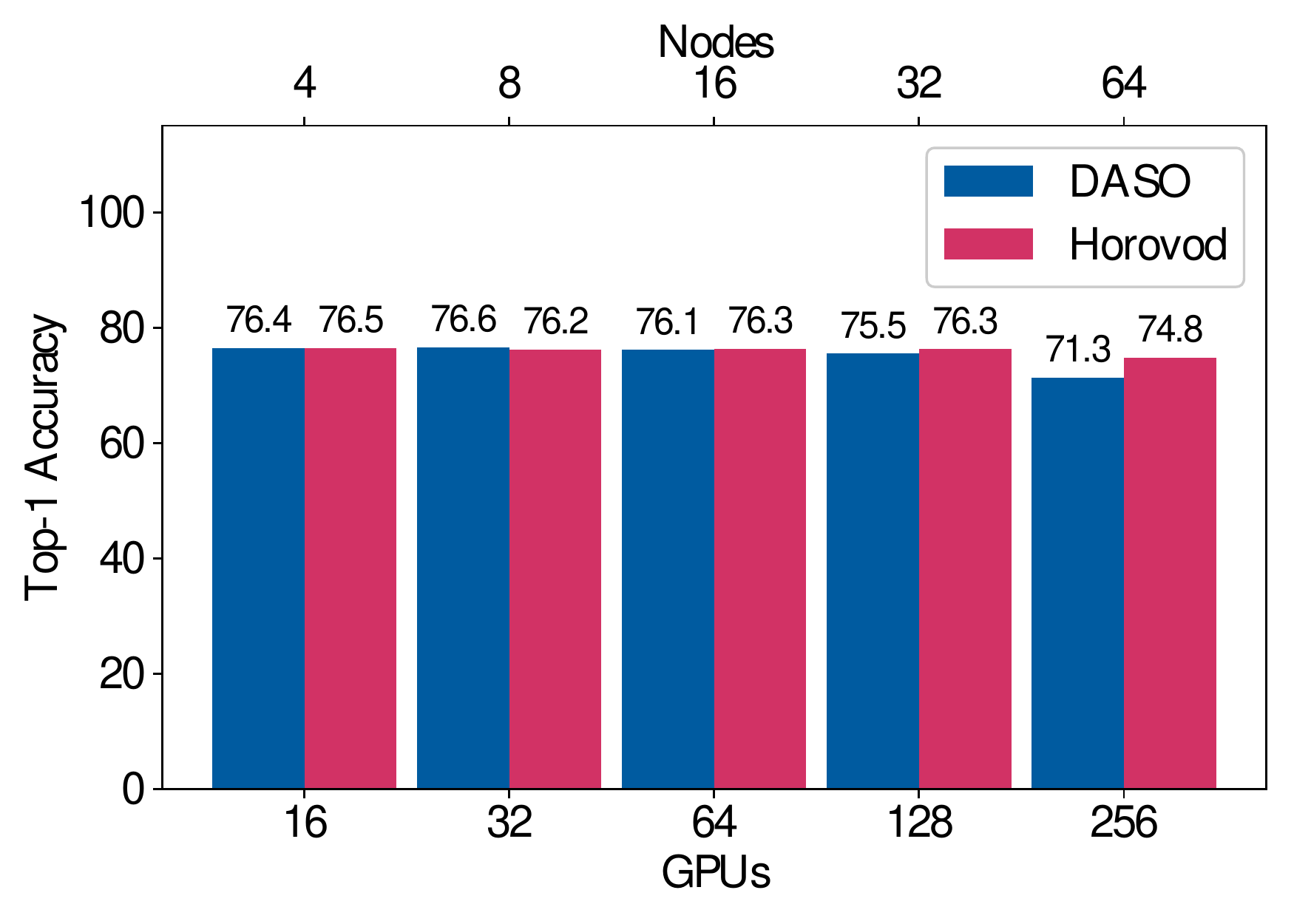} \hfill
    \caption{Top-1 accuracy of ResNet-50 networks on the ImageNet dataset when trained with DASO and Horovod for various node counts, each with 4 GPUs.}
\label{fig:img-acc}
\end{minipage}
\end{figure}

Results of the experiment are shown in \Cref{fig:img-scaling}. Both DASO and Horovod show desirable strong scaling behavior, i.e. a factor of two in GPU number results in the training time being halved. Due to DASO's optimized hierarchical communication scheme and the reduced number of synchronizations, DASO requires up to 25\% less time for training compared to Horovod.

It can further be observed that up to 128 GPUs, DASO and Horovod yield similar levels of accuracy, see \Cref{fig:img-acc}. However, with more than 128 GPUs, both approaches did not exceed 75\% top-1 accuracy. 
This is due to the fact that accuracy starts to a decrease at larger batch sizes in a traditional network unless special allowances are made~\cite{goyal2018accurate}. Since we keep the portion of the distributed batch that is processed on each individual GPU the same, larger GPU counts ultimately result in a larger distributed batch. Hence, accuracy ultimately decreases. For DASO, the effect is more dramatic because completing batches without a global synchronization has a similar effect to increasing the size of the batch.

    

\subsection{Semantic Segmentation -- CityScapes}
\label{sec:cityscapes}

To further evaluate the performance of the DASO method, we conducted experiments on a cutting edge, state-of-the-art network. To this end, a hierarchical multi-scale attention network~\cite{tao2020hierarchical} was trained for semantic segmentation on the CityScapes~\cite{cordts2016cityscapes} dataset. This dataset comprises a collection of images of streets in 50 cities across the world, with 5,000 finely annotated images and 20,000 coarsely annotated images. The network has an HRNet-OCR backbone, a dedicated fully convolutional head, an attention head, and an auxiliary semantic head~\cite{tao2020hierarchical}. 

The quality of semantic segmentation networks is often evaluated based on the intersection over union (IOU)~\cite{rezatofighi2019generalized} score. IOU is defined as the intersection of the correctly predicted annotations with the ground truth annotations, divided by their union. In this work, IOU ranges from 0.0 to 1.0.

The network was trained using the following parameters:
175 epochs; the region mutual information loss~\cite{zhao2019rmiloss} function; a local SGD optimizer with a weight decay of 0.0001, a momentum of 0.9, and an initial learning rate of 0.0125; a learning rate scheduler which decays the learning rate by a factor of 0.75 when the loss is judged to be stable for 5 epochs. 
The number of epochs, loss function, and optimizer settings were determined by the original source
~\cite{tao2020hierarchical}. 
The learning rate scheduler deploys a warm up phase of 5 epochs, in which the learning rate is slowly increased from 0.0 to 0.4, after which it decays as scheduled. For the DASO experiments, the synchronized batch normalization operation is conducted within the node-local process group.

\begin{figure}
\centering
\begin{minipage}{.475\textwidth}
  \centering
    \includegraphics[width=\linewidth]{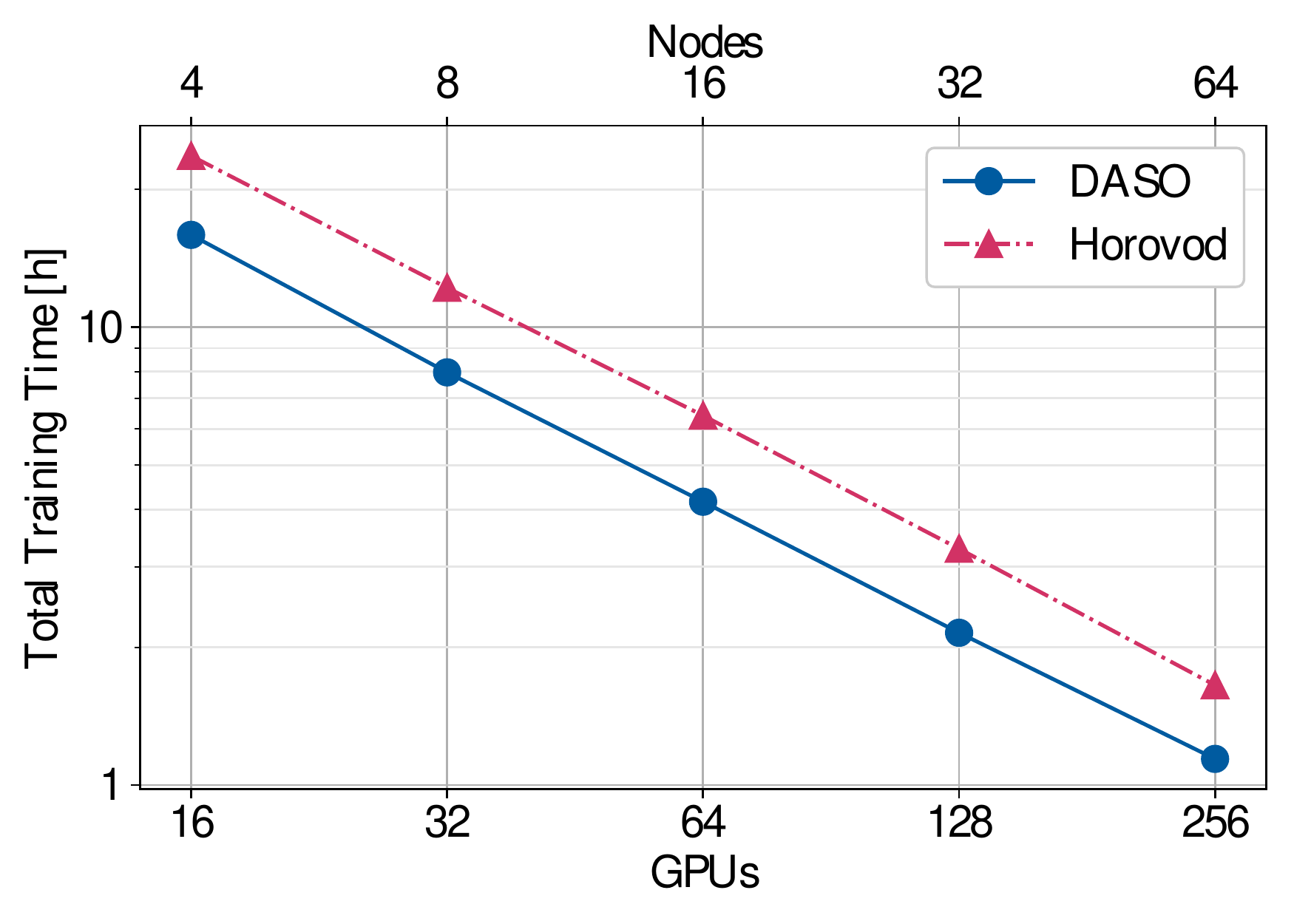} \hfill
    \caption{Training time for the selected hierarchical split level attention network~\cite{tao2020hierarchical} on the CityScapes dataset with DASO and Horovod for increasing node counts, each with 4 GPUs.}
\label{fig:cities-scaling}
\end{minipage}
\hspace{.01\textwidth}
\begin{minipage}{.475\textwidth}
  \centering
    \includegraphics[width=\linewidth]{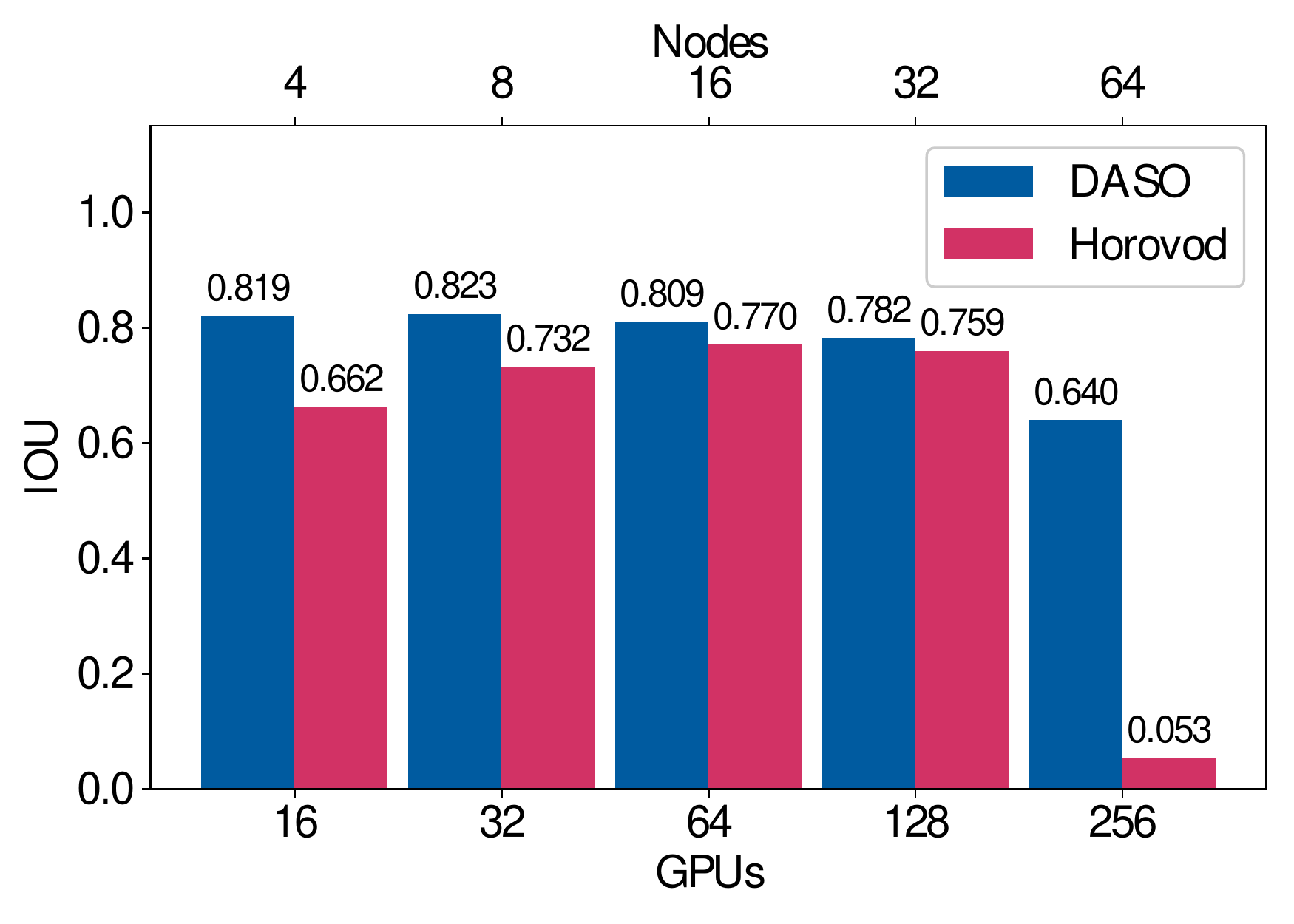} \hfill
    \caption{Maximum IOU of the hierarchical split level attention networks on the CityScapes dataset when trained with DASO and Horovod with various node counts, each with 4 GPUs.}
\label{fig:cities-acc}
\end{minipage}
\end{figure}


In its original publication, the network was trained using supplementary data, whereas the herein presented experiments are performed using only the CityScapes dataset. To determine a baseline accuracy, the original network was trained with four GPUs on a single node using PyTorch's \texttt{DistributedDataParallel} package. This baseline measurement employed a polynomial decay learning rate scheduler, PyTorch's automatic mixed precision training and synchronized batch normalization layers. For more detail, see~\cite{tao2020hierarchical}. The baseline IOU of the original network was found to be $0.8258$. 

During the experiments, we found that for Horovod neither the automatic mixed precision nor the synchronized batch normalization functioned as intended when using the system scheduler software (SLURM~\cite{yoo2003slurm}). Horovod requires usage of its custom scheduler \texttt{horovodrun} to enable full feature functionality. However, this software is not natively available on many computer clusters, including the JUWELS booster supercomputer. Hence, automatic mixed precision was removed and the synchronized batch normalization layers were replaced with local batch normalization layers.

Training times for various node counts are shown in \Cref{fig:cities-scaling}. For up to 128 GPUs, DASO completed the training process in approximately 35\% less time than Horovod, demonstrating the advantage of our approach to fully leverage the systems communication architecture together with asynchronous parameter updates. At higher GPU counts the time savings drop to 30\%, because there are fewer batches per epoch and hence skipping global synchronization operations provides less benefits.


Quality measurements (IOU) are shown in \Cref{fig:cities-acc}. Although there is a very clear difference between Horovod and DASO, neither matches the accuracy of the baseline network. This is due to the naive learning rate scheduler used for training. With a tuned learning rate optimizer the 16, 32, and 64 node configuration should more accurately recreate the results of the baseline network.
At 256 GPUs, training with Horovod did not yield any meaningful results. We hypothesize that this is caused by the lack of a functioning synchronized batch normalization operation in combination with a very large mini-batch. 

\section{Conclusion}
\label{sec:conclusion}

In this work, we have introduced the distributed asynchronous and selective optimization (DASO) method. DASO utilizes a hierarchical communication scheme to fully leverage the communications infrastructure inherent to node-based computer clusters, which often see multiple GPUs per node. By favoring node-local parameter updates, DASO is able to reduce the amount of global communication required for full data parallel network synchronization. Thereby, our approach alleviates the bottleneck of blocking synchronization used in traditional data parallel approaches. We show that, if independent and identically distributed (iid) batches can be reasonably assumed, the global synchronization ubiquitous to the training of DPNNs is not required after each forward-backward pass. 

We evaluated DASO on two common DPNN use-cases: image classification on the ImageNet dataset with ResNet-50, and semantic segmentation on the CityScapes dataset with a cutting edge multi-head attention network architecture.
Our experiments show that DASO can reduce training time by up to 34\% while maintaining similar prediction accuracy when compared to Horovod, the current standard for MPI-based data parallel network training.

At large node counts, DASO and Horovod both suffer a decrease in network accuracy. This is a well-known problem which relates to an increase in the distributed batch size. The effect is more pronounced with DASO due to the reduced number of global synchronization steps. This allows for the identification of where network modifications must be employed to handle very large node counts. We also note that DASO and Horovod will both yield sub-optimal results on datasets for which the iid assumption no longer holds. For those cases, however, data parallel training will be ineffective regardless of the communications scheme. Overall, DASO achieves close-to-optimal accuracies significantly faster than Horovod. Therefore, DASO is optimal for rapid initial training of large networks/datasets, where the training can be further fine-tuned using more traditional methods.

Ultimately, DASO improves the scalability of data parallel neural networks and demonstrates that using more GPUs does not have to be the only solution to speeding up training. With DASO, it is possible to efficiently train large models or process more training data. The beauty of our approach lies in the fact that it is a generic, non-tailored, and easy to implement approach that translates well to any large scale, node-based computer cluster or high-performance computing system. DASO opens the door to redefining data parallel neural network training towards asynchronous, multifaceted optimization approaches.

\section{Acknowledgments}

This work is supported by the Helmholtz Association Initiative and Networking Fund under project number ZT-I-0003, the Helmholtz AI platform grant and the HAICORE@KIT partition. 

\bibliographystyle{unsrt}  
\bibliography{references}

\appendix

\section{Supplementary Materials}

\subsection{Proof of Convergence}

\begin{proof}
The following proof of DASO's global synchronization method is based heavily on the convergence analysis shown by~\cite{Bottou2018OptimizationMF} and will show that the gradients determined with DASO are bounded.

Let $X\subset\mathbb{R}^n$ be a known set, and $f: X\rightarrow\mathbb{R}$ a differentiable, convex, $L$-smooth, and unknown function. Then, the estimator of the stochastic gradient of $f(x)$ is a function $\tilde{g}(x)$ for inputs $x$ determined by the realization of a random variable $\zeta$, such that $\mathbb{E}[\tilde{g}(x;\zeta)]=\nabla f(x:\zeta)$. In the following, $\zeta$ is omitted due to space constraints. The stochastic gradient descent (SGD) algorithm updates a model's state at batch $t+1$, $x_{t+1}$, with the following rule $x_{t+1}=x_t - \eta\tilde{g}(x_t)$, where $\eta$ is the parametric learning rate. 

A commonly used variant of SGD in practice is minibatching for computational efficiency reasons. In minibatch SGD, the true stochastic gradient is approximated by averaging across $m$ input items $x_i$, i.e. $\tilde{G}(x_t)=\frac{1}{m}\sum_{i=1}^{m}\tilde{g}(x_{t, i})$. The model state $x_{t+1}$ for minibatch SGD is
\begin{equation}\label{eq:1}
    x_{t+1} = x_t - \eta \tilde{G}\left( x_t\right)
\end{equation}
where $\tilde{G}\left( x_t\right)$ is an estimator of $\nabla f\left( x_t\right)$.

Let us now consider, that $S$ subsequent update steps are performed. It is possible to write the model state as: 
\begin{equation}\label{eq:2}
    x_{t+S} = x_t - \eta \sum_{i=0}^{S-1} \tilde{G}\left( x_{t+i}\right)
\end{equation}

One of the primary assumptions in SGD is the Lipschitz-continuous objective gradients. This has the effect that:
\begin{align}\label{eq:libshitz}
\begin{split}
    f\left(x_{t+1}\right) - f\left(x_{t}\right) \leq 
    {}& -\eta\nabla f\left(x_{t}\right)^T\mathbb{E}\left[\tilde{g}\left(x_t\right)\right] \\
    & + \frac{1}{2} \eta^2 L\mathbb{E}\left[\left\lVert\tilde{g}\left(x_t\right)\right\rVert_2^2\right]
\end{split}
\end{align}
where the Lipschitz constant, $L$, is greater than zero. \Cref{eq:libshitz} implies that the expected decrease in the objective function, $f(x)$, is bounded above by a set quantity, regardless of how the stochastic gradients arrived at $x_t$~\cite{Bottou2018OptimizationMF}.

In DASO, the local synchronization step is bound via the same assumptions as minibatch SGD outlined in~\cite{Bottou2018OptimizationMF}, so long as the iid assumption is upheld. However, the non-standard global synchronization step used in DASO must be shown to be bound under the same principles. DASO's global synchronization is: 
\begin{equation}
    x^\text{DASO}_{t+S} = \frac{2Sx_{l:t+S-1} + \sum^{P}_{i=1} x^i_{p:t}}{2S + P}
\end{equation}
where the $l$ and $p$ subscripts represent the node-local and global model states, $S$ is the number of local update steps before global synchronization, and $P$ is the number of processes.

Similar to~\Cref{eq:1}, this can also be represented via the locally and globally calculated gradients, 
$\tilde{G_l}\left( x_{l:t}\right)$ and 
$\tilde{G_p}\left( x_{p:t}\right)$ respectively. The global synchronization function in the gradient representation is as follows:
\begin{equation}\label{eq:daso-xts}
    x^\text{DASO}_{t+S} = x_{t} - \alpha
    \left(
    2S \sum^{S-1}_{k=0} \tilde{G}_l \left( x_{l:t + k} \right) +
    \sum^{P}_{i=1} \tilde{G}_p \left( x^i_{p:t} \right)
    \right)
\end{equation}

where $\alpha = \eta / (2S + P)$. 
Using this, \Cref{eq:1}, and the fact that the updates between $t$ and $S$ are local synchronizations which take the form of \Cref{eq:2}, we find that globally calculated gradients are as follows.
\begin{align}\label{eq:daso-grads}
\begin{split}
    \tilde{G}^{\text{DASO}}\left( x_{t+S-1}\right) =
    {}& P\sum^{S-1}_{\beta=0}\tilde{G}_l\left(x_{l:t+S-\beta}\right) \\
    &- 2S\tilde{G}_l\left(x_{l:t+S-1}\right)\\
    &+ \sum^{P}_{i=1} \tilde{G}_p \left( x^i_{p:t} \right)
\end{split}
\end{align}

As all gradient elements in~\Cref{eq:daso-grads} are bound under \Cref{eq:libshitz}, $\tilde{G}^{\text{DASO}}\left( x_{t+S-1}\right)$ is similarly bounded.
\end{proof}

\end{document}